# Object Classification Model Using Ensemble Learning with Gray-Level Co-Occurrence Matrix and Histogram Extraction

Florentina Tatrin Kurniati [1,2], Danny Manongga [1], Eko Sediyono [1], Sri Yulianto Joko Prasetya [1], Roy Rudolf Huizen [3]

[1] Faculty of Information Technology Universitas Kristen Satya Wacana,
Jl. Diponegoro No.52-60, Salatiga, Jawa Tengah 50711
[2] Faculty of Informatics and Computer, Institut Teknologi dan Bisnis STIKOM Bali, Indonesia
Jl. Raya Puputan No.86, Denpasar Bali 80234
[3] Department of Magister Information System, Institut Teknologi dan Bisnis STIKOM Bali, Indonesia
Jl. Raya Puputan No.86, Denpasar Bali 80234

| ARTICLE INFO | ABSTRACT |
|---|---|
| **Article history:**<br>Received July 13, 2023<br>Revised August 13, 2023<br>Published August 22, 2023<br><br>**Keywords:**<br>Classification;<br>Voting Ensemble;<br>Combined Classifier;<br>GLCM;<br>Histogram | In the field of object classification, identification based on object variations is a challenge in itself. Variations include shape, size, color, and texture, these can cause problems in recognizing and distinguishing objects accurately. The purpose of this research is to develop a classification method so that objects can be accurately identified. The proposed classification model uses Voting and Combined Classifier, with Random Forest, K-NN, Decision Tree, SVM, and Naive Bayes classification methods. The test results show that the voting method and Combined Classifier obtain quite good results with each of them, ensemble voting with an accuracy value of 92.4%, 78.6% precision, 95.2% recall, and 86.1% F1-score. While the combined classifier with an accuracy value of 99.3%, a precision of 97.6%, a recall of 100%, and a 98.8% F1-score. Based on the test results, it can be concluded that the use of the Combined Classifier and voting methods is proven to increase the accuracy value. The contribution of this research increases the effectiveness of the Ensemble Learning method, especially the voting ensemble method and the Combined Classifier in increasing the accuracy of object classification in image processing. |

**Corresponding Author**:

Florentina Tatrin Kurniati, Faculty of Information Technology, Universitas Kristen Satya Wacana
Jl. Diponegoro No.52-60, Salatiga, Jawa Tengah 50711
Email: 982022026@student.uksw.edu

## 1.  INTRODUCTION

In the field of classification, grouping based on the nature and characteristics of objects is essential [1], [2]. Object classification is used to differentiate objects in images based on relevant attributes [3], [4]. These problems are found in various fields besides the ability of identification systems based on characteristics. As in research with variations in poses, expressions, and lighting it becomes one of the challenges for identification. Besides that, in the medical field, object identification is a challenge in itself, one of which is for the detection of medical diseases which involves the identification of pathology on medical images. Identification requires extraction, to find out the characteristics of the object, you can use GLCM. In previous research, the system performance showed an accuracy of 0.984, a sensitivity of 0.992, a specificity of 0.968, and a precision of 0.967 with Magnetic Resonance Imaging samples. The classification model using SVM with k-NN obtained results of 94.6% and 91% [5]–[8]. In addition, in the industrial field, object classification for identification is used to group objects based on characteristics. This identification can be used for product identification, sorting, and recognition of an object [9]–[11]. So object identification is a critical study because it requires accurate and effective results. So it is very important to overcome these problems so that the results become reliable [12]–[14]. In object classification, there are several problems that need to be addressed, including variations in the complexity of objects in the dataset, including variations in shape, size, color, texture, and object context. This variation can cause difficulties in recognizing and distinguishing objects accurately [15], [16]. In addition, every single classification model may have specific weaknesses or tend to provide unstable predictive results in some situations [17]–[19].

Based on this, the purpose of this research is to improve the accuracy of object identification. Improvement using several classification methods such as K-Nearest Neighbors (KNN), Random Forest (RF), Support Vector Machine (SVM), Decision Tree, and Naive Bayes (NB). Ensemble Learning methods including Voting and Combined Classifiers, are used to improve reliability [19]–[22]. Voting and Combined Classifier methods use classification, K-Nearest Neighbors (K-NN), Random Forest (RF), Support Vector Machine (SVM), Decision Tree, and Naive Bayes (NB) [23]–[27]. The combination of these methods used to predict and classify allows the combination to improve prediction results. This is based on that each model may have an emphasis on specific characteristics or features that are more relevant in object classification [28]–[31]. By using a combination of classification prediction results from several models, it can improve classification accuracy and reduce the tendency of prediction errors when using a single model. Ensemble Learning methods such as Voting and Combined Classifier can overcome the problem of complexity and variation in object classification. Combining the prediction results from several classification models will improve the accuracy of object classification and provide more reliable results in various image processing and pattern recognition applications [32].

In addition to the Ensemble Learning method, this study also uses GLCM (Gray-Level Co-occurrence Matrix) feature extraction and histograms to obtain relevant characteristics or features from the image. GLCM is an effective method for describing the spatial relationship between pixel intensities in an image. GLCM features namely Contrast, Correlation, Energy, Homogeneity, entropy will be extracted from object images for use in the classification process [33]–[36]. Histogram feature extraction shows the frequency of occurrence of pixel intensity levels. The histogram is represented as a numeric vector that represents the image. Histogram features can be used in classification algorithms to predict unknown image classes. By combining Ensemble Learning and GLCM feature extraction methods and histograms, a more accurate object classification model is produced. The contribution of this research increases the effectiveness of the Ensemble Learning method, by using Voting Ensemble and Combined Classifier to increase reliability.

## 2. METHODS

The object detection method begins with hold pre-processing, as shown in Fig. 1. The object datasets with different sizes are resized to ensure uniform size in order to simplify the extraction process. The extraction process uses the Gray-Level Co-occurrence Matrix (GLCM) and histogram methods [34], [37].

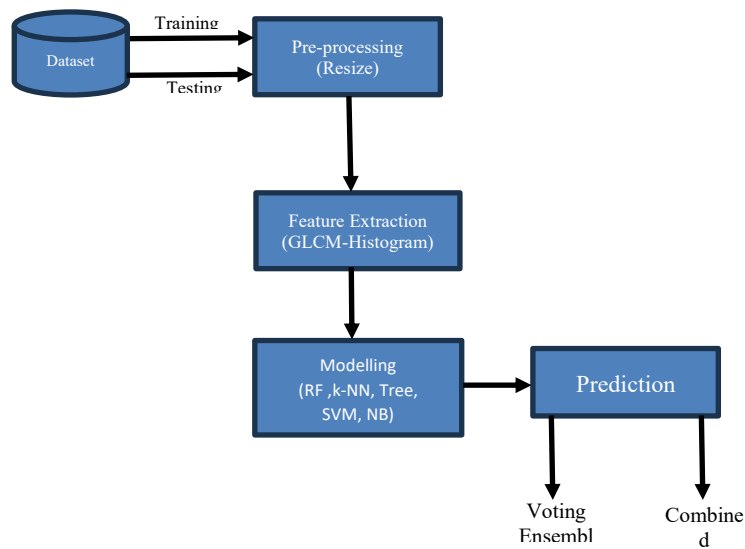

**Fig. 1.** Flowchart Classification Model using Ensemble Learning

Feature extraction uses the GLCM and histogram methods. Features are obtained by analyzing pixel pairs (GLCM) and the distribution of pixels in the sample image (histogram) as shown in Table 1 and Table 2. Features are classified individually using (a). Random Forest (RF) in principle this method uses several tens to hundreds of decision trees for classification, (b). This Support Vector Machine (SVM) method for feature classification uses a hyperplane to maximize the distance of object classes, (c). the k-Nearest Neighbors (k-NN) method classifies object classes based on the majority of nearest neighbor classes, (d). the Naive Bayes method determines class labels based on features, with features not related to each other, (e) while the Decision Tree method uses a tree structure to describe rules and predict classes [38]–[42]. Ensemble voting method by combining the prediction results from various individual classification algorithms. For method (g). combine classifier to improve accuracy based on priority.

$$Energy = \sum_i \sum_j P(i,j)^2 \quad (1)$$

$$Contrast = \sum_i \sum_j (i-j)^2 * P(i,j) \quad (2)$$

$$Homogeneity = \sum_i \sum_j \frac{P(i,j)}{1+(i-j)^2} \quad (3)$$

$$Entropy = -\sum_i \sum_j P(i,j) * log(P(i,j)) \quad (4)$$

$$Correlation = \frac{\sum_i \sum_j [ij * P(i,j)] - \mu_x * \mu_y}{\sigma_x * \sigma_y} \quad (5)$$

Equation (1) Energy measures texture uniformity, calculated by adding the squares of the co-occurrence probabilities of each pair of pixels in the matrix. Equation (2) Contrast measures the variation between neighboring pixels, calculated by taking the difference between the row and column indices, squaring it, and then multiplying it by the probability of co-occurrence. Equation (3) Homogeneity measures the proximity of elements in the co-occurrence matrix. Equation (4) Entropy measures the complexity of the information in an image. Equation (5) Correlation which measures the linear dependence between pixel intensities at positions (i) and (j) the average pixel intensity of each coordinate. Where, i and j are co-occurrence matrix indices, P(i,j) is the element of the co-occurrence matrix at position (i,j), μx and μy are the average row and column weights of the co-occurrence matrix dan σx and σy are the standard deviation of the row and column weights of the co-occurrence matrix.

Table 1. Histogram Pseudo Code

| No | Step Pseudo Code | Pseudo Code |
|---|---|---|
| 1 | Create an array called histogram with size N, initialized with zeros, where N is the number of possible intensity levels in the image | function calculateHistogram(image): |
| 2 | Create an array called histogram with size N, initialized with zeros, where N is the number of possible intensity levels in the image. | histogram = array of size N, initialized with zeros |
| 3 | Get the width and height of the image | width = width of image |
| 4 | Iterate over each pixel in the image using two nested loops, one for the y-coordinate (rows) and one for the x-coordinate (columns). | height = height of image for y from 0 to height-1: |
| 5 | Retrieve the intensity of the pixel at coordinate (x, y) in the image. | for x from 0 to width-1: |
| 6 | Increment the corresponding element in the histogram array by 1 for the found intensity level. | intensity = intensity of pixel at (x, y) in image |
| 7 | Repeat steps 4 and 5 for each pixel in the image. | histogram[intensity] = histogram[intensity] + 1 |
| 8 | Return the calculated histogram | return histogram |

After the dataset pre-processing process is complete, the next step is to perform the extraction using the GLCM (Grey Level Co-occurrence Matrix) and Histogram methods. GLCM is a two-dimensional matrix that describes the relationship between pixel intensities in an image. To describe important information from the matrix by calculating the energy, which describes the intensity of the pixels scattered in the image matrix. For contrast measure the significant difference in pixel intensity in the matrix. Meanwhile, homogeneity is calculated to measure the extent to which the pixel intensities are similar in the matrix. Meanwhile, entropy is used to describe the level of disorder or complexity in the image matrix. The next feature is a correlation to measure the linear relationship between the pixel intensities in the matrix. The next feature extraction is the histogram, which is the analysis of the pixel intensity distribution in an image. The histogram is used to determine the distribution of pixel intensity across a range of possible values. The histogram extraction process uses the stages of changing the grayscale image and calculating the histogram at the pixel gray level. determine the range of minimum and maximum intensity values in the image, the range is divided into several intervals and adjusted to the needs of the analysis. Each pixel in the image is placed into the appropriate bin based on its intensity. This is done by comparing the pixel intensity values with predetermined bin interval limits. The number of pixels in the bin is counted and represents the distribution of the frequency or intensity of the pixels in each value interval. The histogram formula is shown in Table 1. Improved predictions using the Voting Ensemble and Combined Classifier methods. The Voting Ensemble method collects predictions from each model and votes based on a majority. Whereas Combined Classifier is more dynamic, a model with unknown label predictions, then predictions from other models will replace it as shown in Table 2.

**Table 2.** Voting Ensemble and Combined Classifier

| No | Pseudo Code | Algorithm |
|---|---|---|
| 1 | FUNCTION VotingEnsemble(RF_predict, SVM_predict, kNN_predict, NB_predict, DT_predict)<br>FOR i = 1 TO length(RF_predic)<br>Create a list: predict_list = [RF_predict[i], SVM_predict[i], kNN_predict[i], NB_predict[i], DT_predict[i]]<br>vote = MostFrequentLabel(predict_list)<br>ensemble_predict[i] = vote<br>END FOR<br>RETURN ensemble_predict<br>END FUNCTION | Voting Ensemble |
| 2 | FUNCTION CombinedEnsemble(RF_predict, SVM_predict, kNN_predict, NB_predict, DT_predict)<br>FOR i = 1 TO length(RF_predict)<br>IF RF_predict[i] != "unknown"<br>ensemble_predict[i] = RF_predict[i]<br>ELSE IF SVM_predict[i] != "unknown"<br>ensemble_predict[i] = SVM_predict[i]<br>ELSE IF kNN_predict[i] != "unknown"<br>ensemble_predict[i] = kNN_predictions[i]<br>ELSE IF NB_predict[i] != "unknown"<br>ensemble_predict[i] = NB_predictions[i]<br>ELSE<br>ensemble_predict[i] = DT_predictions[i]<br>END IF<br>END FOR<br>RETURN ensemble_predict<br>END FUNCTION | Combined Classifier |
| | // Main execution<br>TRAIN each model (RF, SVM, k-NN, NB, Decision Tree) using training data<br>PREDICT with each model using test data<br>voting_results = VotingEnsemble(RF_predict, SVM_predict, kNN_predict, NB_predict, DT_predict)<br>combined_results = CombinedEnsemble(RF_predict, SVM_predict, kNN_predict, NB_predict, DT_predict) | |

## 3. RESULTS AND DISCUSSION

The confusion matrix is used to test and find out the accuracy of the tested model for classification models with Random Forest (RF) Fig. 2(a), accurate prediction results in all classes.

In Fig. 2(b), for the K-Nearest Neighbors (KNN) model, the results of classifying several samples experienced prediction errors in a number of classes. In Fig. 2(c) Decision Tree model also shows a similar pattern, although with a slightly lower error rate than KNN. Meanwhile, Fig. 2(d) Supports Vector Machine (SVM) model also faces challenges in classifying the different classes, especially in predicting the first and second classes with a significant number of errors. Meanwhile, Fig. 2(e) Naive Bayes (NB) model also shows performance similar to SVM, with a significant error rate in the first and second-class predictions. The test results show that the Random Forest classification model has an accuracy of 99.09%, this method also shows superiority in precision and recall with 99.28% and 98.96% respectively. This indicates that the Random Forest not only classifies most of the samples correctly but also exhibits a good balance in minimizing errors. While SVM showed the lowest performance with 43.47% accuracy, with precision reaching 43.52%, and 41.48% recall, indicating that SVM is often mistaken in identification. The k-NN and Tree methods show average performance with an accuracy of 76.13% and 79.73%, respectively. Both have a balance between precision and recall, indicating that they have a relatively balanced error rate for positive and negative classifications. Meanwhile, Naive Bayes has an accuracy of 50.90%, with a precision of 56.55% but a lower recall of 46.09%. The test results show that Random Forest shows the best and most consistent performance in all evaluation metrics.

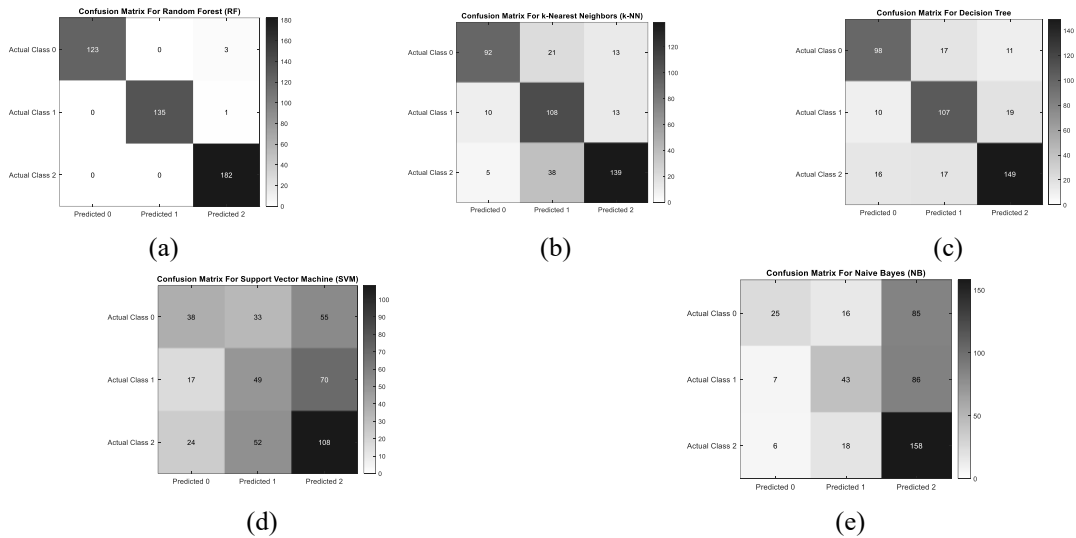

**Fig. 2.** Confusion Matrix (a) Random Forest (b) k-NN (c) Decesion Tree
(d) SVM (e) Naive Bayes

Based on the independent classification, RF, k-NN, SVM, Tree, and NB are used in the Voting Ensemble and Combined Classifier models (Fig. 3(a) and Fig. 3(b)). The result is that the Combined Classifier method has an accuracy of 98.88%, a precision of 99.01%, a recall of 98.72%, and an F1-score of 98.86%. Meanwhile, the Voting Ensemble accuracy was 87.39%, the average precision was 88.42%, the average recall was 86.24%, and the F1 score was 86.96%. These results show that the Combined Classifier is able to classify better than the Voting Ensemble model.

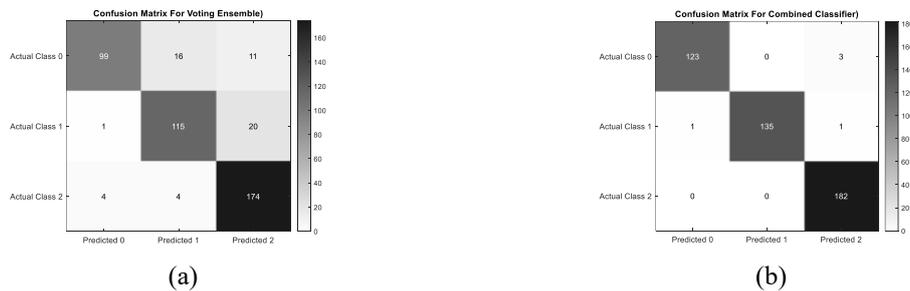

**Fig. 3.** Confusion Matrix (a) Voting Ensemble (b) Combined Classifier

The Confusion matrix in Voting Ensemble has a high level of accuracy in predicting Class 0, class 2, and Class 3, with 99 correct predictions for Class 0, 135 correct predictions for Class 1, and 174 correct predictions for Class 2. However, this model has a little difficulty in predicting Class 2, with 16 prediction errors in Class 1 and 11 prediction errors in Class 2. Overall, Voting Ensemble shows good performance with a high degree of accuracy. Meanwhile, the confusion matrix on the Combined Classifier shows almost identical results to the Voting Ensemble, with a high degree of accuracy in all classes. This model predicts class 1 and class 3 perfectly, with 123 correct predictions for class 1 and 182 correct predictions for class 3. Similar to the Voting

Ensemble, this model has little difficulty predicting Class 2, with only 1 prediction error to Class 1 and 3 prediction errors to Class 3. Overall, the Combined Classifier also shows very good performance with a high degree of accuracy.

The test results in the form of a bar chart are shown in Fig. 4 prediction results 4(a). Class-0, 4(b). Class-1 and 4(c). Class-2, it can be seen that the SVM (Support Vector Machine) and NB (Naive Bayes) models have a lower prediction success rate for all classes compared to the other models. In addition, RF (Random Forest), VE (Voting Ensemble), and CC (Combined Classifier) seem to do a very good job predicting all classes, with bar heights reaching almost 100%. Models that are not accurate in predicting are class 1: SVM and NB models, class 2 SVM and NB models, and class 3 SVM and NB models. In general, SVM and NB models appear to be the least accurate in predicting all classes.

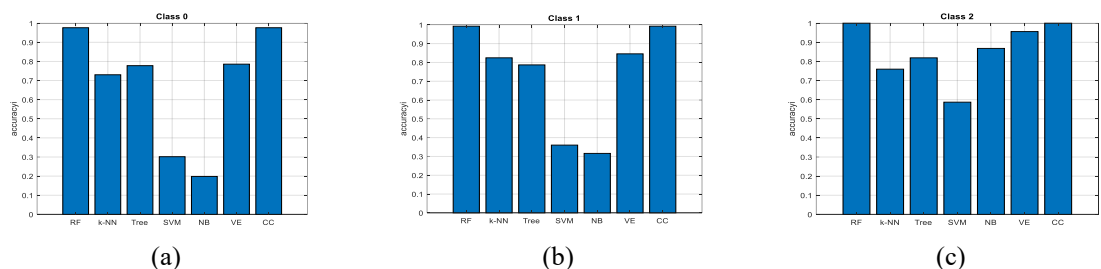

(a) (b) (c)
**Fig. 4.** Class Prediction Results

**Table 3.** Classification Test Results

| Classifier | Accuracy | Precision | Recall | F1 Score |
|---|---|---|---|---|
| Random Forest (RF) | 0.993 | 0.976 | 1.000 | 0.988 |
| k-Nearest Neighbors (k-NN) | 0.871 | 0.730 | 0.852 | 0.786 |
| Decision Tree (Tree) | 0.868 | 0.778 | 0.790 | 0.784 |
| Support Vector Machine (SVM) | 0.599 | 0.302 | 0.481 | 0.371 |
| Naive Bayes (NB) | 0.665 | 0.198 | 0.658 | 0.305 |
| Voting Ensemble (VE) | 0.924 | 0.786 | 0.952 | 0.861 |
| Combined Classifier (CC) | 0.993 | 0.976 | 1.000 | 0.988 |

The highest accuracy, shown in Table 3 value is in the Random Forest (RF) and Combined Classifier classification models, with an accuracy of 0.993. Meanwhile, the model with the lowest accuracy is the Support Vector Machine (SVM) with an accuracy of 0.599. Precision is the ratio of True Positives divided by the total number of True Positives and False Positives. This shows how often the model is right when it predicts the positive class. RF and Combined Classifiers also have the highest precision, with a value of 0.976. The model with the lowest precision is Naive Bayes (NB), with a value of 0.198. Recall or Sensitivity is the ratio of True Positives divided by the total number of True Positives and False Negatives which indicates how often the model finds a positive class when it is actually a positive class. RF and Combined Classifier have the highest recall, with a value of 1,000. The model with the lowest recall is SVM, with a value of 0.481. F1 Score is the harmonic mean of precision and recall. The F1 score tries to strike a balance between precision and recall. RF and Combined Classifier have the highest F1 score, with a value of 0.988. These results indicate that the stretch model is capable of solving problems related to object variations [5]–[8], [43].

## 4. CONCLUSION

The classification models evaluated were Random Forest (RF), K-Nearest Neighbors (KNN), Decision Tree, Support Vector Machine (SVM), and Naive Bayes (NB), the results of Random Forest (RF) have high accuracy. Meanwhile, the SVM and NB models experienced difficulty in classification, with SVM recording the lowest accuracy of 43.47%. Even though the k-NN model and Decision Tree have moderate performance, the precision and gain values are balanced. The classification model is used in the Voting Ensemble model with an accuracy of 87.39%, while the Combined Classifier shows superiority with an accuracy of 98.88%, precision of 99.01%, recall of 98.72%, and F1 score of 98.86%. The Voting Ensemble and Combined Classifier models open opportunities for wider development for models with low accuracy.

## REFERENCES

[1] X. X. Bao, C. Zhao, S. S. Bao, J. S. Rao, Z. Y. Yang, and X. G. Li, "Recognition of necrotic regions in MRI images of chronic spinal cord injury based on superpixel," *Comput. Methods Programs Biomed.*, vol. 228, p. 107252, 2023,

https://doi.org/10.1016/j.cmpb.2022.107252.
[2] S. Salem Ghahfarrokhi, H. Khodadadi, H. Ghadiri, and F. Fattahi, "Malignant melanoma diagnosis applying a machine learning method based on the combination of nonlinear and texture features," *Biomed. Signal Process. Control*, vol. 80, no. P1, p. 104300, 2023, https://doi.org/10.1016/j.bspc.2022.104300.
[3] S. Rooj, A. Routray, and M. K. Mandal, "Feature based analysis of thermal images for emotion recognition," *Eng. Appl. Artif. Intell.*, vol. 120, p. 105809, 2023, https://doi.org/10.1016/j.engappai.2022.105809.
[4] Y. Park and J. M. Guldmann, "Measuring continuous landscape patterns with Gray-Level Co-Occurrence Matrix (GLCM) indices: An alternative to patch metrics?," *Ecol. Indic.*, vol. 109, no. June 2019, p. 105802, 2020, https://doi.org/10.1016/j.ecolind.2019.105802.
[5] M. Sood, S. Jain, and J. Dogra, "Classification and Pathologic Diagnosis of Gliomas in MR Brain Images," *Procedia Comput. Sci.*, vol. 218, pp. 706–717, 2023, https://doi.org/10.1016/j.procs.2023.01.051.
[6] O. Attallah and D. A. Ragab, "Auto-MyIn: Automatic diagnosis of myocardial infarction via multiple GLCMs, CNNs, and SVMs," *Biomed. Signal Process. Control*, vol. 80, no. P1, p. 104273, 2023, https://doi.org/10.1016/j.bspc.2022.104273.
[7] D. Rammurthy and P. K. Mahesh, "Whale Harris hawks optimization based deep learning classifier for brain tumor detection using MRI images," *J. King Saud Univ. - Comput. Inf. Sci.*, vol. 34, no. 6, pp. 3259–3272, 2022, https://doi.org/10.1016/j.jksuci.2020.08.006.
[8] D. Rahadiyan, S. Hartati, Wahyono, and A. P. Nugroho, "Feature aggregation for nutrient deficiency identification in chili based on machine learning," *Artif. Intell. Agric.*, vol. 8, pp. 77–90, 2023, https://doi.org/10.1016/j.aiia.2023.04.001.
[9] D. Caceres-Hernandez *et al.*, "Recent advances in automatic feature detection and classification of fruits including with a special emphasis on Watermelon (Citrillus lanatus): A review," *Neurocomputing*, vol. 526, pp. 62–79, 2023, https://doi.org/10.1016/j.neucom.2023.01.005.
[10] Y. M. Tang, W. Li, W. T. Kuo, and C. K. M. Lee, "Real-time Mixed Reality ( MR ) and Artificial Intelligence ( AI ) Object Recognition Integration for Digital Twin in," *Internet of Things*, p. 100753, 2023, https://doi.org/10.1016/j.iot.2023.100753.
[11] Y. Li, H. S. Abdel-Khalik, A. Al Rashdan, and J. Farber, "Feature extraction for subtle anomaly detection using semi-supervised learning," *Ann. Nucl. Energy*, vol. 181, p. 109503, 2023, https://doi.org/10.1016/j.anucene.2022.109503.
[12] M. Redhya and K. Sathesh Kumar, "Refining PD classification through ensemble bionic machine learning architecture with adaptive threshold based image denoising," *Biomed. Signal Process. Control*, vol. 85, p. 104870, 2023, https://doi.org/10.1016/j.bspc.2023.104870.
[13] G. Wu *et al.*, "Early identification of strawberry leaves disease utilizing hyperspectral imaging combing with spectral features, multiple vegetation indices and textural features," *Comput. Electron. Agric.*, vol. 204, 2023, https://doi.org/10.1016/j.compag.2022.107553.
[14] G. Saleem, M. Akhtar, N. Ahmed, and W. S. Qureshi, "Automated analysis of visual leaf shape features for plant classification," *Comput. Electron. Agric.*, vol. 157, pp. 270–280, 2019, https://doi.org/10.1016/j.compag.2018.12.038.
[15] A. S. Nasution, A. Alvin, A. T. Siregar, and M. S. Sinaga, "KNN Algorithm for Identification of Tomato Disease Based on Image Segmentation Using Enhanced K-Means Clustering," *Kinet. Game Technol. Inf. Syst. Comput. Network, Comput. Electron. Control*, vol. 4, no. 3, 2022, https://doi.org/10.22219/kinetik.v7i3.1486.
[16] A. N. Vasquez, D. María Ballesteros, and D. Renza, "Machine learning applied to speaker verification of fake voice recordings," *2019 22nd Symp. Image, Signal Process. Artif. Vision, STSIVA 2019 - Conf. Proc.*, no. 1, 2019, https://doi.org/10.1109/STSIVA.2019.8730273.
[17] A. Nanda, R. C. Barik, and S. Bakshi, "SSO-RBNN driven brain tumor classification with Saliency-K-means segmentation technique," *Biomed. Signal Process. Control*, vol. 81, p. 104356, 2023, https://doi.org/10.1016/j.bspc.2022.104356.
[18] L. Cui, W. Zhang, R. Zhang, H. Zhai, X. Zhang, and X. Xie, "Researches on the novel methodology of traffic flow prediction based on similarity," *2011 Int. Conf. Soft Comput. Pattern Recognit.*, pp. 296–300, 2011, https://doi.org/10.1109/SoCPaR.2011.6089259.
[19] S. Q. Dong *et al.*, "How to improve machine learning models for lithofacies identification by practical and novel ensemble strategy and principles," *Pet. Sci.*, vol. 20, no. 2, pp. 733–752, 2023, https://doi.org/10.1016/j.petsci.2022.09.006.
[20] O. El Alaoui and A. Idri, "Predicting the potential distribution of wheatear birds using stacked generalization-based ensembles," *Ecol. Inform.*, vol. 75, p. 102084, 2023, https://doi.org/10.1016/j.ecoinf.2023.102084.
[21] A. Ghasemieh, A. Lloyed, P. Bahrami, P. Vajar, and R. Kashef, "A novel machine learning model with Stacking Ensemble Learner for predicting emergency readmission of heart-disease patients," *Decis. Anal. J.*, vol. 7, p. 100242, 2023, https://doi.org/10.1016/j.dajour.2023.100242.
[22] S. Wu, M. Liang, X. Wang, and Q. Chen, "VGbel: An exploration of ensemble learning incorporating non-Euclidean structural representation for time series classification," *Expert Syst. Appl.*, vol. 224, p. 119942, 2023, https://doi.org/10.1016/j.eswa.2023.119942.
[23] S. Vidivelli and S. Sathiya Devi, "Breast cancer detection model using fuzzy entropy segmentation and ensemble classification," *Biomed. Signal Process. Control*, vol. 80, no. P1, p. 104236, 2023, https://doi.org/10.1016/j.bspc.2022.104236.
[24] A. M. Vommi and T. K. Battula, "A hybrid filter-wrapper feature selection using Fuzzy KNN based on Bonferroni

mean for medical datasets classification: A COVID-19 case study," *Expert Syst. Appl.*, vol. 218, p. 119612, 2023, https://doi.org/10.1016/j.eswa.2023.119612.
[25] V. Nemade and V. Fegade, "Machine Learning Techniques for Breast Cancer Prediction," *Procedia Comput. Sci.*, vol. 218, no. 2022, pp. 1314–1320, 2023, https://doi.org/10.1016/j.procs.2023.01.110.
[26] S. Shakil, D. Arora, and T. Zaidi, "Feature based classification of voice based biometric data through Machine learning algorithm," *Mater. Today Proc.*, vol. 51, pp. 240–247, 2021, https://doi.org/10.1016/j.matpr.2021.05.261.
[27] S. Shakil, D. Arora, and T. Zaidi, "An optimal method for identification of finger vein using supervised learning," *Meas. Sensors*, vol. 25, p. 100583, 2023, https://doi.org/10.1016/j.measen.2022.100583.
[28] M. Kumar, S. Gupta, and N. Mohan, "A computational approach for printed document forensics using SURF and ORB features," *Soft Comput.*, vol. 24, no. 17, pp. 13197–13208, 2020, https://doi.org/10.1007/s00500-020-04733-x.
[29] A. Ali, M. Hamraz, N. Gul, D. M. Khan, S. Aldahmani, and Z. Khan, "A k nearest neighbour ensemble via extended neighbourhood rule and feature subsets," *Pattern Recognit.*, vol. 142, p. 109641, 2023, https://doi.org/10.1016/j.patcog.2023.109641.
[30] R. Islam, M. I. Sayed, S. Saha, M. J. Hossain, and M. A. Masud, "Android malware classification using optimum feature selection and ensemble machine learning," *Internet Things Cyber-Physical Syst.*, vol. 3, pp. 100–111, 2023, https://doi.org/10.1016/j.iotcps.2023.03.001.
[31] E. Kesriklioğlu, E. Oktay, and A. Karaaslan, "Predicting total household energy expenditures using ensemble learning methods," *Energy*, vol. 276, 2023, https://doi.org/10.1016/j.energy.2023.127581.
[32] R. Soni and B. Mehta, "Diagnosis and prognosis of incipient faults and insulation status for asset management of power transformer using fuzzy logic controller & fuzzy clustering means," *Electr. Power Syst. Res.*, vol. 220, p. 109256, 2023, https://doi.org/10.1016/j.epsr.2023.109256.
[33] Y. Chen *et al.*, "Experimental investigation on the fracture surface features of heat-treated red sandstone containing fissure under constant amplitude low cycle impact using 3D digital reconstruction," *Eng. Fract. Mech.*, vol. 277, p. 109002, 2023, https://doi.org/10.1016/j.engfracmech.2022.109002.
[34] G. Kostka, L. Steinacker, and M. Meckel, "Under big brother's watchful eye: Cross-country attitudes toward facial recognition technology," *Gov. Inf. Q.*, vol. 40, no. 1, 2023, https://doi.org/10.1016/j.giq.2022.101761.
[35] V. Sathiyamoorthi, A. K. Ilavarasi, K. Murugeswari, S. Thouheed Ahmed, B. Aruna Devi, and M. Kalipindi, "A deep convolutional neural network based computer aided diagnosis system for the prediction of Alzheimer's disease in MRI images," *Meas. J. Int. Meas. Confed.*, vol. 171, p. 108838, 2021, https://doi.org/10.1016/j.measurement.2020.108838.
[36] S. S. Gudadhe, A. D. Thakare, and D. Oliva, "Classification of intracranial hemorrhage CT images based on texture analysis using ensemble-based machine learning algorithms: A comparative study," *Biomed. Signal Process. Control*, vol. 84, p. 104832, 2023, https://doi.org/10.1016/j.bspc.2023.104832.
[37] E. Shaw, S. Snow, and C. Timmermann, "From mouthpiece of an emerging specialty to voice for high-quality research: the first 100 years of the British Journal of Anaesthesia," *Br. J. Anaesth.*, pp. 1–8, 2023, https://doi.org/10.1016/j.bja.2023.04.007.
[38] X. Wang, W. Cai, and M. Wang, "A novel approach for biometric recognition based on ECG feature vectors," *Biomed. Signal Process. Control*, vol. 86, no. PA, p. 104922, 2023, https://doi.org/10.1016/j.bspc.2023.104922.
[39] Z. Ullah, L. Qi, A. Hasan, and M. Asim, "Improved Deep CNN-based Two Stream Super Resolution and Hybrid Deep Model-based Facial Emotion Recognition," *Eng. Appl. Artif. Intell.*, vol. 116, , p. 105486, 2022, https://doi.org/10.1016/j.engappai.2022.105486.
[40] G. Guven, U. Guz, and H. Gürkan, "A novel biometric identification system based on fingertip electrocardiogram and speech signals," *Digit. Signal Process. A Rev. J.*, vol. 121, p. 103306, 2022, https://doi.org/10.1016/j.dsp.2021.103306.
[41] S. Khan and M. Narvekar, "Novel fusion of color balancing and superpixel based approach for detection of tomato plant diseases in natural complex environment," *J. King Saud Univ. - Comput. Inf. Sci.*, vol. 34, no. 6, pp. 3506–3516, 2022, https://doi.org/10.1016/j.jksuci.2020.09.006.
[42] H. Fauzi, C. Erika, S. Sa'adiah, and F. Oscandar, 'Classification of Gender Individual Identification Using Local Binary Pattern on Palatine Rugae Image', *J. Ilm. Tek. Elektro Komput. dan Inform.*, vol. 8, no. 3, p. 422, 2022, https://doi.org/10.26555/jiteki.v8i3.23636.
[43] A. Ali, M. Hamraz, N. Gul, D. M. Khan, S. Aldahmani, and Z. Khan, 'A k nearest neighbour ensemble via extended neighbourhood rule and feature subsets', *Pattern Recognit.*, vol. 142, p. 109641, 2023, https://doi.org/10.1016/j.patcog.2023.109641.

## BIOGRAPHY OF AUTHORS

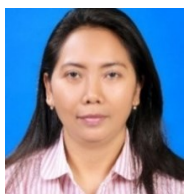

**Florentina Tatrin Kurniati** received her Master in Informatics Engineering from Atma Jaya Yogyakarta University (UAJY), Yogyakarta, Indonesia (2015), and is currently pursuing a doctoral program in computer science at Satya Wacana Christian University. Since 2008 he has been a lecturer and researcher at the faculty informatics and computers, Institut Teknologi dan Bisnis STIKOM Bali, Indonesia. He is interested in adaptive noise cancellation, pattern recognition, Object Identification, and digital forensics.

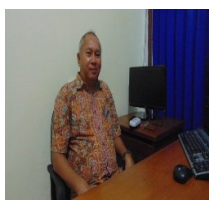
**Daniel HF Manongga** received an Engineer degree in electronics from Satya Wacana Christian University Salatiga in 1981, Master of Science in information technology from Queen Mary College-University of London in 1989, and Doctor of Philosophy in information systems, artificial intelligence and management science from Queen Mary College-University of London in 1996. His research areas include artificial intelligence, cloud computing, and the semantic web. Currently active as a teacher at the Faculty of Information Technology, Satya Wacana Christian University, Salatiga.

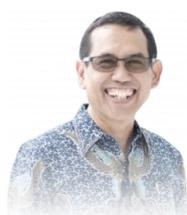
**Eko Sediyono,** completed his professorship in 2008. He is a member of the IEEE Computer Society #41605422. He completed his undergraduate studies in 1985 at the Bogor Agricultural Institute majoring in Statistics. Received a master's degree in 1994 from the University of Indonesia and a doctorate in 2006 from the University of Indonesia majoring in Computer Science. Currently working as a lecturer at Satya Wacana Christian University Salatiga and serving as Deputy Chancellor for Research, Innovation and Entrepreneurship at Satya Wacana Christian University-Indonesia. Research interests in Data Science, Algorithms and Image Processing. Completed professorship in 2008.

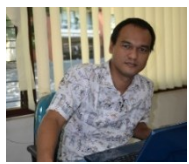
**Sri Yulianto Joko Prasetyo**, completed his doctorate degree on the Doctorate Program of Computer Science, Science Faculty of Gadjah Mada University in 2013. He has been active on research since 2008 until now on the Spatial Data Processing and Remote Sensing. He has published his papers on international journals Scopus Indexed.

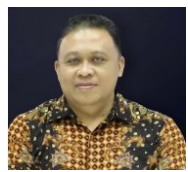
**Roy Rudolf Huizen,** Graduated with Doctor of Computer Science (2018) from Universitas Gadjah Mada (UGM) Yogyakarta, Indonesia. Lecturer and researcher at the Department of Magister Information System at the Institut Teknologi dan Bisnis STIKOM Bali, with research interests in the fields of Object Identification, Signal Processing, Cyber Security Forensics and Artificial Intelligence.